\documentclass[10pt,journal,compsoc]{IEEEtran}

\usepackage[nocompress]{cite}

\usepackage{epsfig}
\usepackage{graphicx}
\usepackage{amsmath}
\usepackage{amssymb}

\usepackage[accsupp]{axessibility}  % Improves PDF readability for those with disabilities.

\usepackage{graphicx}
\usepackage{amsmath}
\usepackage{amssymb}
\usepackage{booktabs}
\usepackage{algorithm}
\usepackage{algorithmic}

\usepackage[pagebackref,breaklinks,colorlinks]{hyperref}

\usepackage{multirow}

\usepackage{enumitem}
\usepackage{multirow}
\usepackage{stfloats}
\usepackage{tabularx}
\usepackage{threeparttable}
\usepackage[usenames,dvipsnames]{xcolor}

\usepackage{bm}

\newlength\savedwidth

\definecolor{darkergreen}{RGB}{21, 152, 56}
\definecolor{red2}{RGB}{252, 54, 65}

\definecolor{blacktext}{RGB}{0, 0, 0}

\usepackage{xcolor}
\definecolor{yl_color}{RGB}{128, 255, 0}

\definecolor{blue2}{RGB}{20, 54, 254}
\usepackage{bbding}

\usepackage[american]{babel}
\usepackage{microtype}
\usepackage{cleveref}

\usepackage{subcaption} 
\usepackage{cutwin}

\begin{document}

\title{TextMonkey: An OCR-Free Large Multimodal Model for Understanding Document}
% \title{Backbone. }

\author{
Yuliang Liu, IEEE Member,
Biao Yang,
Qiang Liu, 
Zhang Li,
Zhiyin Ma,
Shuo Zhang, \\
Xiang Bai*, IEEE Senior Member
\IEEEcompsocitemizethanks{
\IEEEcompsocthanksitem Y. Liu, B. Yang, Z. Li, Z. Ma, S. Zhang, and X. Bai are with the School of Artificial Intelligence and Automation, Huazhong University of Science and Technology, Wuhan, 430074, China (email: \{ylliu, hust\_byang, xbai\}@hust.edu.cn). Q. Liu are with Kingsoft, Wuhan, 430074, China}
% <-this % stops a space
\thanks{Y. Liu and B. Yang contributed Equally. Corresponding author: X. Bai.}
}

\IEEEtitleabstractindextext{%
\begin{abstract}
We present TextMonkey, a large multimodal model (LMM) tailored for text-centric tasks. Our approach introduces enhancement across several dimensions: By adopting Shifted Window Attention with zero-initialization, we achieve cross-window connectivity at higher input resolutions and stabilize early training; We hypothesize that images may contain redundant tokens, and by using similarity to filter out significant tokens, we can not only streamline the token length but also enhance the model's performance. Moreover, by expanding our model's capabilities to encompass text spotting and grounding, and incorporating positional information into responses, we enhance interpretability. It also learns to perform screenshot tasks through finetuning. Evaluation on 12 benchmarks shows notable improvements: 5.2\% in Scene Text-Centric tasks (including STVQA, TextVQA, and OCRVQA), 6.9\% in Document-Oriented tasks (such as DocVQA, InfoVQA, ChartVQA, DeepForm, Kleister Charity, and WikiTableQuestions), and 2.8\% in Key Information Extraction tasks (comprising FUNSD, SROIE, and POIE). It outperforms in scene text spotting with a 10.9\% increase and sets a new standard on OCRBench, a comprehensive benchmark consisting of 29 OCR-related assessments, with a score of 561, surpassing previous open-sourced large multimodal models for document understanding. Code will be released at \url{https://github.com/Yuliang-Liu/Monkey}.

\end{abstract}

}

\maketitle

\

%
% \tableofcontents
% \clearpage
%

% \emph{i.e.}EEdisplaynontitleabstractindextext
% \emph{i.e.}EEpeerreviewmaketitle

\IEEEraisesectionheading{\section{Introduction}
\label{sec:introduction}}

  \IEEEPARstart{E}{xtracting} key information from a variety of sources, including documents like tables, forms, and invoices, as well as text in the wild is crucial for industries and academic research, aiming to automate and refine document-based and scene-text workflows. 
  This field requires text detection and recognition in both document images and real-world scenes, language comprehension, and the integration of vision and language. 
  
  Many early methods~\cite{tang2023udop,huang2020layoutlmv3} attempt to address the task using a two-stage approach: 1) Detect and recognize the text using external systems; 2) Document understanding based on the fusion of text results and images. However, the individual step of text reading in the processing pipeline may lead to the accumulation of errors. Moreover, relying on off-the-shelf OCR Models/APIs
 (OCR-Models) introduces additional engineering complexity, limits the connection between the text and its surrounding context, and can potentially increase computational costs. To alleviate the drawbacks of external systems before understanding, OCR-Free solutions~\cite{kim2022donut,lee2023pix2struct} have attracted increasing attention recently. 

    \begin{figure}[htbp]
        \centering
    \includegraphics[width=0.45\textwidth]{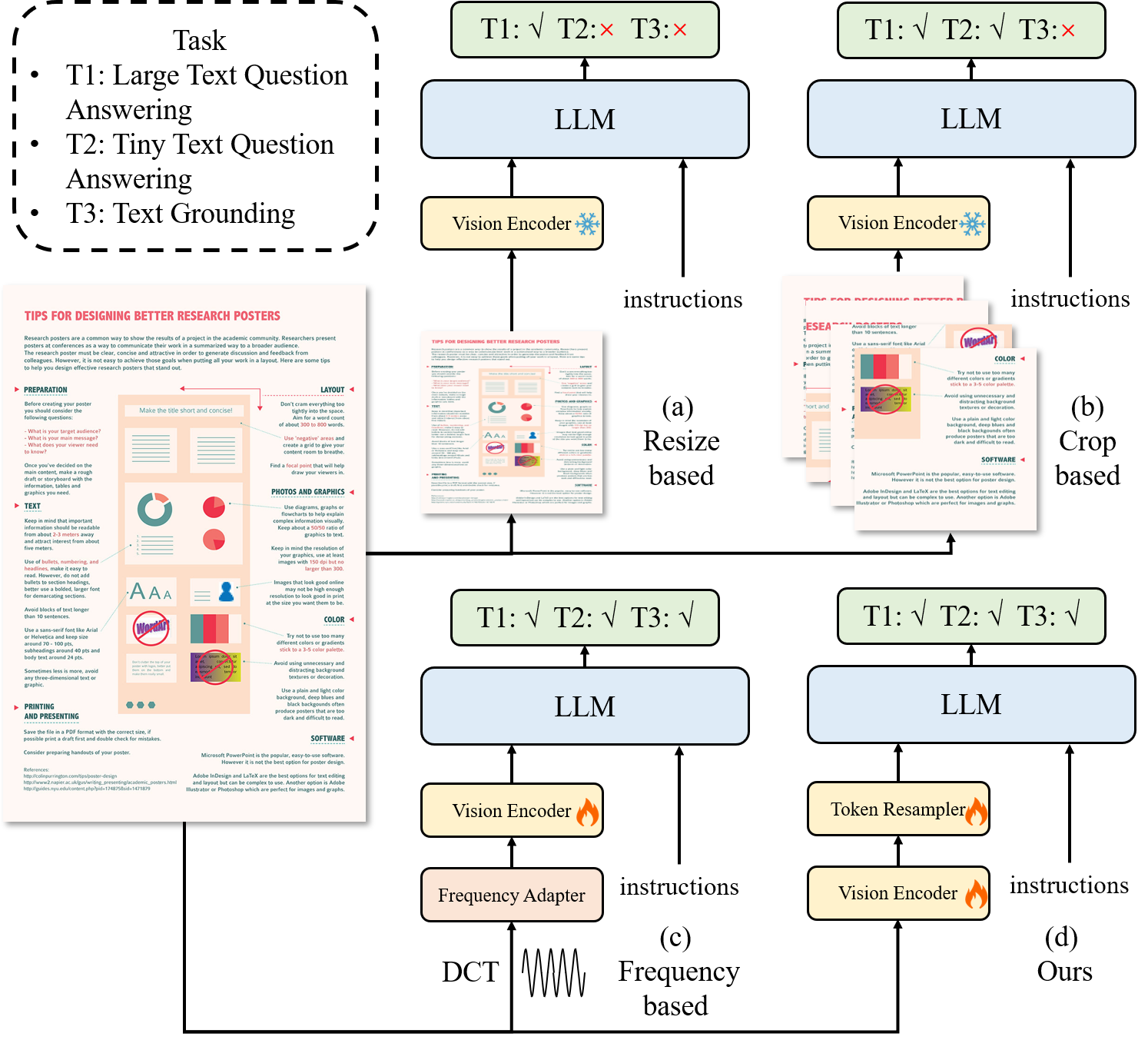}
        \caption{Comparisons to the existing pipelines for document understanding. Compared to  (a) Resize based methods, (b) Crop based methods, and (c) frequency based methods, our model can efficiently process high-resolution text-related images with various tasks.
        }
        \label{fig:diff}
    \end{figure}

    The field of large multimodal models (LMMs)~\cite{liu2023llava,zhu2023minigpt4} is advancing rapidly due to its powerful ability to handle diverse types of data. However, they still have limitations when it comes to addressing text-related tasks.
    As depicted in Fig.~\ref{fig:diff} (a), several methods, including LLaVAR~\cite{zhang2023llavar}, UniDoc~\cite{feng2023unidoc}, TGDoc~\cite{wang2023TGDoc}, and mPLUG-DocOwl~\cite{ye2023mplug-docowl} heavily rely on a pre-trained CLIP~\cite{radford2021clip} for visual encoding.
    Nevertheless, these encoders have input resolutions of 224 or 336, which are insufficient to meet the demands of documents containing numerous small texts~\cite{liu2023hidden}. Therefore, they can only recognize large text and struggle with small text in images.
    To address the limitations of tiny text, UReaer~\cite{ye2023UReader} and Monkey~\cite{li2023monkey} take a cropping strategy to expand the input resolution, as shown in Fig.~\ref{fig:diff} (b). However, this crop strategy may inadvertently split related words, resulting in semantic incoherence. For example, the word "Backup" may be divided into "Back" and "up," making it impossible to restore its original meaning even after fusion has been performed. Besides, the spatial separation caused by this splitting also makes it challenging to handle text position-related tasks, such as text grounding.
    As shown in Fig.~\ref{fig:diff} (c), DocPedia~\cite{feng2023docpedia} directly processes visual input in the frequency domain rather than the pixel space. Due to the characteristics of the frequency domain, it can quickly expand the resolution without losing information. However, due to the transformation of the feature space, it is difficult to leverage existing pretrained models, increasing the demand for training resources.

    We want to inherit the efficient image resolution scaling feature of Monkey~\cite{li2023monkey} but address the missing cross-window context for the documents mentioned above. For this purpose, we introduce TextMonkey, as shown in Fig.~\ref{fig:diff} (d).  TextMonkey utilizes a Split Module that divides high-resolution images into window patches using a sliding window method. Inspired by~\cite{liu2021swin}, we treat every self-attention layer in the CLIP as self-attention in non-overlapped windows. To introduce cross-window relationships while maintaining efficient computation, we use Shifted Window Attention with zero-initialization to build cross-window connections.
    This approach allows us to maintain the training data distribution for the encoder and deal with high-resolution document images while reducing the computational cost of training from scratch. On the other hand, the utilization of the Split Module still poses a significant challenge as it leads to a notable increase in token length. We have observed that there are numerous repetitive image features that align with the language space, similar to certain repeated elements in the language itself. Thus, we propose a token resampler to compress these features while keeping as many of the most important features as possible. We employ important tokens as queries and the original features as key-value pairs, facilitating the reaggregation of features. 
    On the basis of reducing the number of tokens,  our module can also significantly improve the performance compared to random queries.

    On the other hand, due to the self-explanatory nature of the text, in most cases, humans are able to locate the position of the answer itself. To alleviate the issue of hallucination in large language models further, we require the model not only to provide accurate answers but also to locate specific visual evidence supporting its response. We also introduce a variety of text-related tasks to deepen the connection between text information and visual information, such as text spotting and text grounding. Besides, incorporating positional cues into the answers can further enhance the model's reliability and interpretability. 
    
     We summarize the advantages of our method as follows:
     \begin{itemize}
         \item \textbf{Enhancing cross-window relations}. 
         We adopt Shfited Window Attention to successfully incorporate cross-window connectivity while expanding the input resolutions. Besides, we introduce zero initialization in the Shifted Window Attention mechanism, enabling the model to avoid drastic modifications to early training.
         \item \textbf{Token compression}. 
         We show enlarging resolution results in some redundant tokens. By using similarity as a criterion, we are able to find significant tokens that serve as queries for the token resampler. This module not only reduces the token length but also improves the model's performance. Additionally, it significantly improves the performance compared to the use of random queries.
         \item \textbf{Support text grounding}. 
         We expand our scope to include tasks beyond text-based question answering, encompassing reading text, text spotting, and text grounding. Additionally, we found that incorporating positional information into the answers can improve the model's interpretability. TextMonkey can also be finetuned to understand the command of screen-shot clicking. 
         \item We evaluated TextMonkey's performance across 12 recognized benchmarks, observing significant improvements in several areas. Firstly, in scene text-centric tasks such as STVQA, TextVQA, and OCRVQA, TextMonkey achieved a 5.2\% increase in performance. For document-oriented tasks, including DocVQA, InfoVQA, ChartVQA, DeepForm, Kleister Charity, and WikiTableQuestions, it showed a 6.9\% improvement. In the domain of key information extraction tasks, like FUNSD, SROIE, and POIE, we noted a 2.8\% uplift. Particularly notable was its performance in scene text spotting tasks (Total-Text, CTW1500, and ICDAR 2015) focused on transcription accuracy, where it improved by 10.9\%. Additionally, TextMonkey set a new high score of 561 on OCRBench, a comprehensive benchmark encompassing 29 OCR-related evaluations, significantly surpassing the performance of previous open-source, large-scale multimodal models designed for document understanding. This achievement underscores TextMonkey's effectiveness and advances in the field of document analysis and understanding.
     \end{itemize}

\section{Related works}
\label{sec:rela}
Models designed to comprehend images with text information can be broadly categorized into two types: OCR-Model-Driven and OCR-Free methods.

\subsection{OCR-Model-Driven Methods}
OCR-Model-Driven methods use OCR tools to acquire text and bounding box information. Subsequently, they rely on the models to integrate text, layout, and visual data. Meanwhile, diverse pre-training tasks are devised to enhance cross-modal alignment between visual and text inputs.  
StrucTexT~\cite{li2021structext} pays attention to the fine-grained semantic information and global layout information within the image in the design of pre-training tasks. Based on layout knowledge enhancement technology, ERNIE-Layout~\cite{peng2022ernie} innovatively proposes two self-supervised pre-training tasks: reading order prediction and fine-grained image-text matching.
The LayoutLM~\cite{xu2020layoutlm,xu2020layoutlmv2,huang2020layoutlmv3} series continuously improves by integrating pre-trained text, layout, and visual features and introducing a unified model architecture and pre-training goals. This enhances the model's performance in various document understanding tasks and simplifies overall design. UDOP~\cite{tang2023udop} unifies vision, text, and layout through VTL Transformer and a unified generative pre-training task. 
Wukong-reader~\cite{bai2023wukong-reader} proposes the Textline-Region Contrastive Learning and specially crafted pre-training tasks to extract fine-grained text line information. 
DocFormerv2~\cite{appalaraju2023docformerv2} designs an asymmetric pre-training method and a simplified visual branch for visual document understanding. DocLLM~\cite{wang2023docllm}  focuses exclusively on position information to incorporate the spatial layout structure, using a decomposed attention mechanism to build a cross-alignment between text and spatial modalities.

While advancements have been achieved, OCR-Model-Driven methods depend on text extraction from external systems, which necessitates increased computational resources and extends processing durations. Additionally, these models may inherit OCR inaccuracies, presenting challenges to document understanding and analysis tasks.

\subsection{OCR-Free Methods}
OCR-Free methods do not require off-the-shelf OCR engines/APIs.
Donut~\cite{kim2022donut} first proposes an end-to-end training method based on a Transformer without OCR. 
Dessurt~\cite{davis2022dessurt}, based on an architecture similar to Donut, incorporates two-way cross-attention and employs distinct pre-training methods. 
Pix2Struct~\cite{lee2023pix2struct} is pre-trained by learning to parse masked screenshots of web pages into simplified HTML, introducing a variable-resolution input representation and a more flexible way to integrate language and visual input. 
StrucTexTv2~\cite{yu2021structextv2} introduces a novel self-supervised pre-training framework, employing text region-level document image masking to learn end-to-end visual-textual representations. 

Although these methods do not require OCR tool limitations, they still need fine-tuning for specific tasks. In the fast-growing era of Multi-Modal Large Language Models (MLLMs), some models are explicitly trained on visual text understanding datasets and fine-tuned with instructions. 
LLaVAR~\cite{zhang2023llavar}, mPLUG-DocOwl~\cite{ye2023mplug-docowl} and UniDoc~\cite{feng2023unidoc} create novel instruction-following datasets to enhance the tuning process and improve the comprehension of text-rich images. Additional efforts have been undertaken to capture more intricate textual details. UReader~\cite{ye2023UReader} designs a shape-adaptive cropping module that utilizes a frozen low-resolution visual encoder to process high-resolution images. DocPedia~\cite{feng2023docpedia} processes visual input in the frequency domain rather than pixel space to process higher-resolution images with limited visual tokens.
By training a visual vocabulary on a large amount of data, Vary~\cite{wei2023vary} expands its resolution and achieves impressive results. 
Recently, TGDoc~\cite{wang2023TGDoc} uses text-grounding to enhance document understanding, suggesting that textual grounding can improve the model's ability to interpret textual content, thereby enhancing its understanding of images rich in textual information. 
    \begin{figure*}[tbp]
        \centering
        \includegraphics[width=0.9\textwidth]{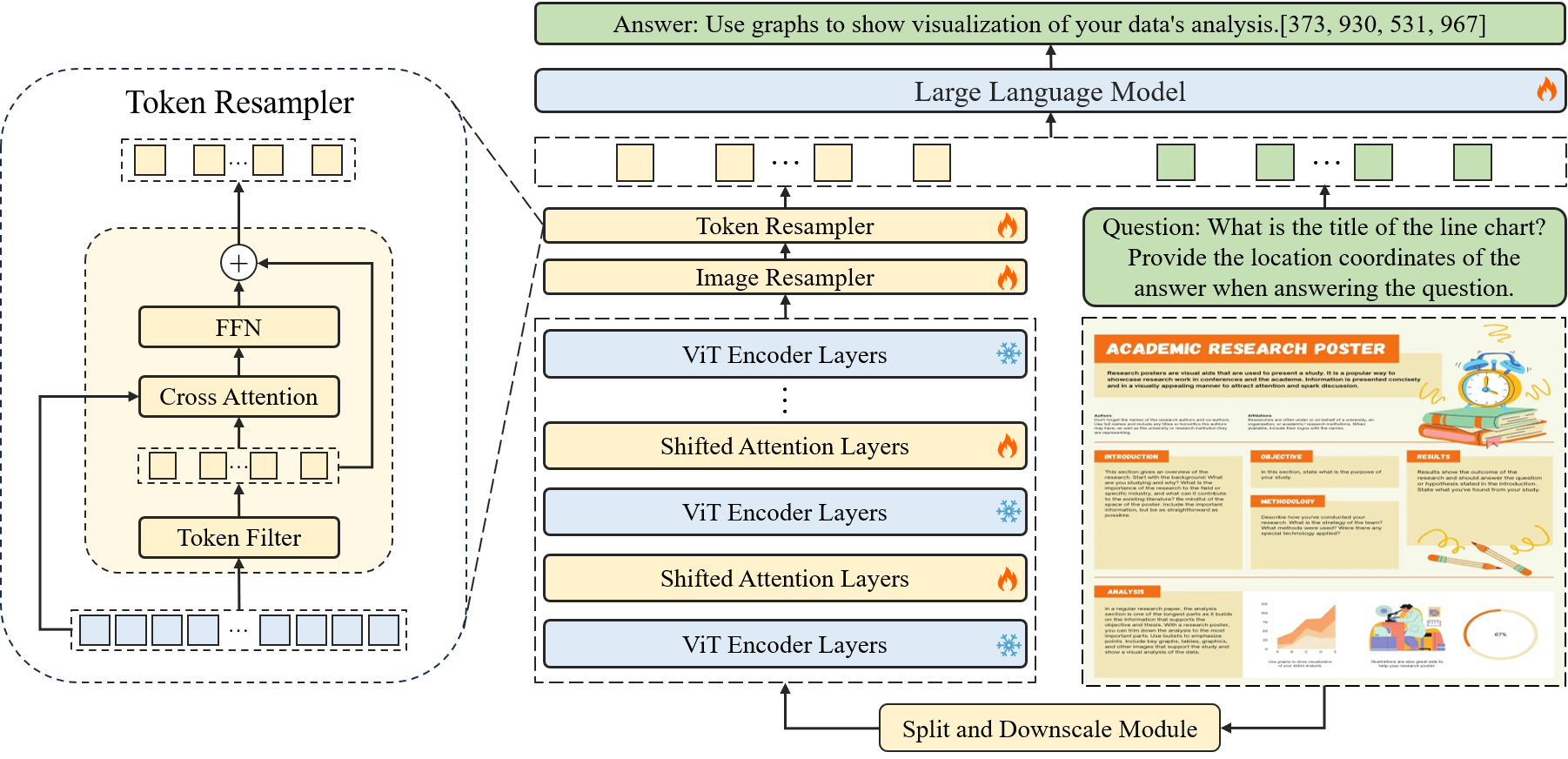}
        \caption{An overview of the TextMonkey.  It enables the enhancement of resolution with limited training resources while preserving cross-window information and reducing redundant tokens introduced by resolution enhancement. Besides, through various data and pretext prompts, TextMonkey has been equipped with the ability to handle multiple tasks.
        }
        \label{fig:method}
    \end{figure*}
    
\section{Methodology}
    \label{sec:method}
    
    The method presented in Fig.~\ref{fig:method} begins by dividing the input image into non-overlapping patches using a sliding window module, with each patch sized at 448x448 pixels. These patches are further subdivided into smaller patches of 14x14 pixels, each considered as a token. Utilizing Transformer blocks that inherit from the pre-trained CLIP model, we process these tokens on each window patch separately. To establish connections among various window patches, Shifted Window Attention is integrated at specific intervals within the Transformer blocks. To generate a hierarchical representation, the input image is resized to 448x448 and fed into CLIP to extract a global feature, as suggested by~\cite{li2023monkey}. This global feature, alongside features from sub-images, is then processed by a shared image resampler to align with the language domain. Then, a Token Resampler is employed to further minimize redundancy in the language space by compressing the length of tokens. Ultimately, these processed features, combined with the input question, are analyzed by a Large Language Model (LLM) to produce the required answers.

\subsection{Shifted Window Attention}
    Previous studies have underscored the significance of input resolution for precise document understanding \cite{feng2023docpedia,liu2023hidden}. To enhance training efficiency, recent methods \cite{li2023monkey,ye2023UReader} have adopted a sliding window technique to enhance image resolution. While effective in analyzing natural scenes due to their localized content, this strategy may lead to the fragmentation of connected text in document analysis, disrupting semantic continuity. Additionally, the spatial disjunction poses challenges for tasks that rely on text positioning, such as text grounding.
    
    To alleviate the issue mentioned above, we adopt Shifted Window Attention~\cite{liu2021swin} to augment the CLIP model's visual processing capabilities.
    Specifically, for an input image $I \in \mathbb{R}^{H\times W \times 3}$, our approach slices the image into non-overlapping windows. This slice is achieved using a sliding window $W \in \mathbb{R}^{H_v\times W_v}$, where $H_v$ and $W_v$ indicate the window's size. Within each window, we independently apply a transformer block from the CLIP architecture, which initially does not account for cross-window relationships. To incorporate interactions between different windows and enhance the model's contextual understanding of the image, we adopt the Shifted Window Attention (SWA) mechanism. 
    As mentioned in ~\cite{liu2021swin}, the sliding window is cyclic-shifting toward the top-left direction, resulting in new windows. The self-attention computation by a masking mechanism, which limits self-attention computation to within new windows.

    To achieve smoother training initialization, we have made modifications to the shifted window attention by allowing them to start learning from zero initialization,  avoiding excessive transformation of early features during the initial stages.
    In particular, we modify the regular initialization in MLP to zero initialization to achieve smoother training, inspired by~\cite{hu2021lora}: 
    \begin{equation}
        x = \textbf{BA}\hat{x},
    \end{equation}
    where \textbf{B} and \textbf{A} refer to the weight of two linear layers.
    We use a random Gaussian initialization for \textbf{A} and zero initialization for \textbf{B}. This approach ensures that the image encoder's parameters remain stable in the initial phase, facilitating a smoother training experience.
    
    \begin{figure}[!t]
        \centering
        \includegraphics[width=1.0\linewidth]{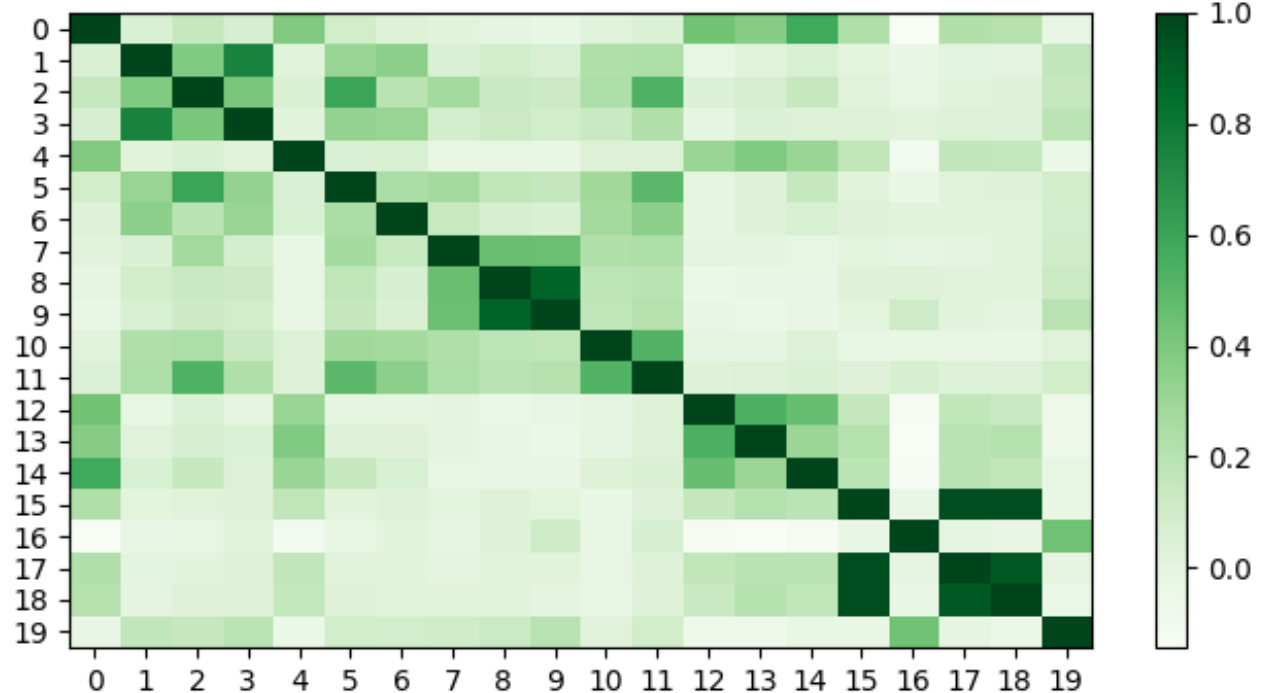}
        \caption{Image token similarity comparisons. We randomly select 20 ordered tokens from image tokens and use cosine similarity as the metric for measuring similarity. 
        }
        \label{fig:hot}
    \end{figure}

\subsection{Image Resampler} 
To reduce the redundancy in image features initially, we inherited the image resampler from Qwen-VL~\cite{bai2023qwen-vl}, which is using upon every window. The module employs a set of trainable parameters as query vectors and utilizes the image features from the visual encoder as keys and values for cross-attention operations. This process helps compress the visual feature sequence to a fixed length of 256. Furthermore, to preserve positional information crucial for fine-grained image comprehension, 2D absolute positional encodings are integrated into the query-key pairs of the cross-attention mechanism.
\subsection{Token Resampler}
    As the resolution increases, the number of tokens also significantly increases, using the slide window mechanism. However, due to limitations in the input length of some language models and training time constraints, reducing the number of tokens becomes necessary.
    In common visual scenarios, the previous method~\cite{bolya2022token} has demonstrated the feasibility of merging token approaches.

\begin{figure}[!t]
        \centering
        \includegraphics[width=0.85\linewidth]{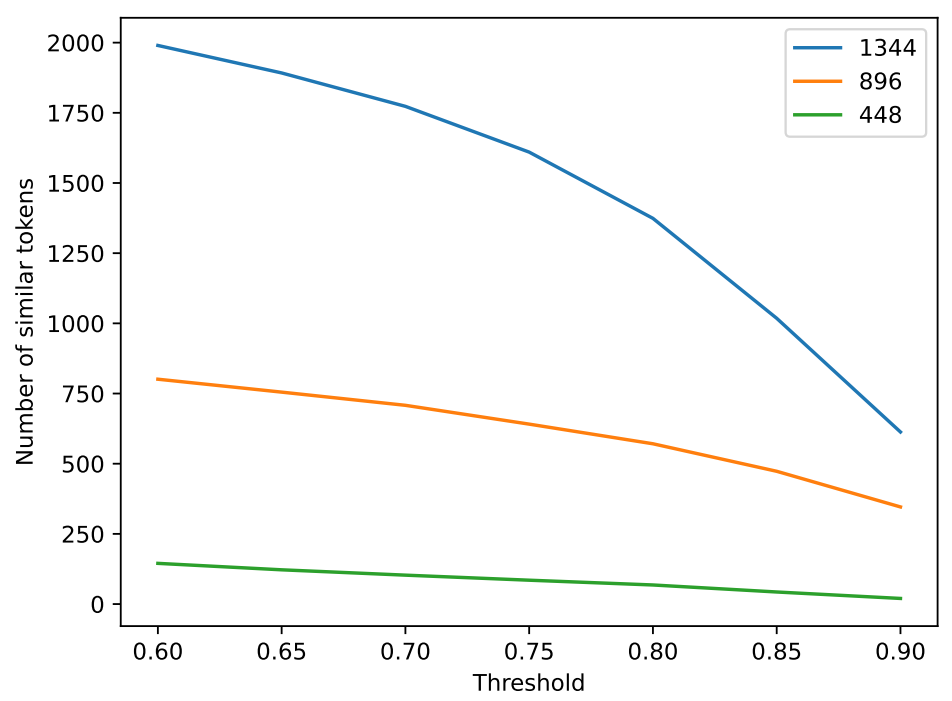}
        \caption{Quantitative analysis on specific redundant tokens. Using the maximum cosine similarity between each token and other tokens as a criterion for identifying redundant tokens, we plotted the threshold on the x-axis and the number of redundant tokens at different resolutions on the y-axis.
        }
        \label{fig:sim}
    \end{figure}

    For natural language, redundant information could be repeated linguistic elements. Assuming that by expanding the resolution of the image, redundant visual information will exist. When determining the similarity between two linguistic elements, we often measure their embeddings' similarity. To assess the redundancy of image features, we measure the similarity of image tokens already mapped to the language space. We randomly select 20 ordered features after the image resampler and compare pairwise similarities using cosine similarity, as shown in Fig.~\ref{fig:hot}. Through the comparison of image tokens' similarity, we can observe a pattern where many image tokens exhibit multiple similar tokens.
    Furthermore, we quantitatively compared the redundancy of tokens at different resolutions, as shown in Fig.~\ref{fig:sim}. 
    Empirically, we selected a threshold value of 0.8 as the similarity threshold,  At resolutions of 448, 896, and 1334, we observed 68/256 (26.6\%), 571/1024 (55.8\%), and 1373/2304 (59.5\%) redundant tokens, respectively.
    As presented in Fig.~\ref{fig:sim}, with an increase in resolution, there is a higher occurrence of repeated tokens.
    This validates our hypothesis that while expanding the resolution can achieve clearer visibility, it also introduces some redundant features.   
    
    However, how can we identify important tokens and eliminate redundant ones? We have observed that certain tokens are highly unique and lack closely similar counterparts, such as the fifth token in Fig.~\ref{fig:hot}. This suggests that this token is distinct. We hypothesize that these tokens carry crucial and distinctive information, which is further validated in subsequent experiments. Therefore, we utilize similarity as a metric to identify significant tokens. 
    
    Hence, we propose a Token Resampler to compress redundant tokens, as shown in the left part of Fig.~\ref{fig:method}. As shown in Algor. \ref{algor:1}, we utilize a token filter algorithm to select the most valuable tokens.

\begin{algorithm}[ht]
  \caption{\small\label{filter}Token Filter Implementation}\label{algor:1}
  \begin{algorithmic}[1]
    \REQUIRE tokens $ \in \mathbb{R}^{L \times D}$, r (remain token numbers) \\
    CMX (calculate max similarity)
    \STATE  importances = []\\
    \STATE  for token in tokens:\\
    \STATE \quad max\_similarity = CMX(token, other\_tokens) \\
    \STATE \quad importances.append(1-max\_similarity) \\
    \STATE top\_tokens = select\_top\_tokens(tokens, importances, r) \\
    \STATE  sorted\_tokens = sort\_by\_original\_order(top\_tokens) \\
    \STATE Return sorted\_tokens.
  \end{algorithmic}
\end{algorithm}

    To avoid information loss caused by directly discarding other tokens, 
    we utilize important tokens as queries and employ cross-attention to further aggregate all the features. Based on the reduction of the token count, our module can also significantly improve the performance compared to random queries.

\subsection{Position-Related Task}
   To alleviate the issue of hallucinations in Large Language Models (LLMs), where they can produce incorrect responses not related to the provided image, we aim to enhance their capability to analyze and incorporate visual information into their replies. 
    Considering that answers to text-based tasks are often found within the image itself, we anticipate that the large model will not only produce precise responses but also identify the particular visual proof that underpins its answer.

    Moreover, we have undertaken modifications to existing question-answering datasets. Specifically, we have found the positions with the majority of answers in the images. These positional cues have been extracted and seamlessly integrated into the answers themselves. To preserve the original capability of direct dialogue, we have also retained the original question-answering task.

    For better perception of the spatial positions of the text, it requires the model to have a strong spatial understanding. Building upon the aforementioned model designs, we add additional training tasks to improve the model's perception of text positions, such as text spotting and reading text. Specific tasks and prompts are shown in Tab.~\ref{tab:prompt}. 
    To guarantee a strong connection between text and location data, we strictly maintain their alignment, ensuring that text information always comes before any associated location details.

    To standardize images of different ratios, we use a scale of (0, 1000) to represent positional information. Therefore, in an image with resolutions of  ($H_r\times W_r$), the text coordinates (x, y) will be normalized to $[ (x/H_r*1000)]$, and the same applies to y. The restoration process involves the inverse operation.

\begin{table}[]
\centering
\caption{Prompts for a variety of tasks.}
\label{tab:prompt}
\begin{tabular}{c|c}
\toprule
Type & Prompt \\ \midrule
Read All Text & Read all the text in the image. \\ \midrule
Text Spotting & OCR with grounding: \\ \midrule
Original Tasks & \{Question\}.  Answer: \\ \midrule
Position of text & \textless ref\textgreater text\textless/ref\textgreater \\ \midrule
\multirow{2}{*}{Text Recognition} & \textless ref\textgreater This\textless/ref\textgreater \\
 & \textless box\textgreater(x1,y1),(x2,y2)\textless/box\textgreater is \\ \midrule
\multirow{2}{*}{VQA Grounding} & \{Question\}. Provide the location coordinates \\ 
 & of the answer when answering the question. \\ \bottomrule
\end{tabular}
    
\end{table}
    \subsection{Dataset Construction}
    \label{subsec:dc}

    During our training process, we solely utilize open-source data and apply various task-specific augmentations to different datasets.
    By integrating various datasets and employing different instructions for different tasks, we enhance the model's learning ability and training efficiency. For scene text scenario, we select COCOText~\cite{veit2016cocotext}, TextOCR~\cite{singh2021textocr}, HierText~\cite{long2022towards}, TextVQA~\cite{singh2019towards}, and MLT~\cite{nayef2019icdar2019} for training. For document images, we select IIT-CDIP~\cite{lewis2006building}, DocVQA~\cite{mathew2021docvqa}, ChartQA~\cite{masry2022chartqa}, InfoVQA~\cite{infovqa}, DeepForm~\cite{deepform}, Kleister Charity (KLC)~\cite{stanislawek2021kleister}, and WikiTableQuestions (WTQ)~\cite{pasupat2015compositional}.
     To accelerate the training speed, we have transformed single-image question answering into multi-turn image-based question answering, significantly improving the utilization of image features, following the successful approach introduced in LLaVA~\cite{liu2023llava}.
    The details of our training data are shown in Tab.~\ref{tab:data}. We have a total of 409.1k pairs of dialogue data and 2.1M question-answering pairs in our dataset to train our model.

    To further strengthen the model's ability to handle structured text, we fine-tune one epoch on TextMonkey with structured data to enhance its structured capabilities, resulting in TextMonkey\dag. The fine-tuning data primarily consisted of 5\% of the data from the previous stage, as well as a portion of structured data, including documents, tables, and charts. The structured data images are also sourced from publicly available datasets and are generated using their structure information.
    Therefore, we have a total of 55.7k of data in structured data.

 \begin{table}[]
    \centering
        \caption{Details of the training data, derived entirely from publicly available datasets.}
    \begin{tabular}{c|c|c}
    \toprule 
    Task                                    & Dataset          & Samples \\ \midrule

    \multirow{5}{*}{Scene Text} & COCOText ~\cite{veit2016cocotext}         & 16.2k     \\
                                            & TextOCR ~\cite{singh2021textocr}          & 42.7k    \\
                                            & HierText ~\cite{long2022towards}          & 59.9k    \\
                                            & TextVQA ~\cite{singh2019towards}          & 42.6k    \\
                                            & MLT ~\cite{nayef2019icdar2019}            & 11.6k     \\ \midrule 
    \multirow{8}{*}{Document}                & IIT-CDIP ~\cite{lewis2006building}          & 154.6k    \\
                                            & DocVQA ~\cite{mathew2021docvqa}          & 22.7k    \\
                                            & ChartQA ~\cite{masry2022chartqa}         & 36.6k     \\
                                            & InfoVQA ~\cite{infovqa}         & 10.9k     \\
                                            & DeepForm ~\cite{deepform}        & 1.4k      \\
                                            & KLC ~\cite{stanislawek2021kleister}             & 5.2k     \\
                                            & WTQ ~\cite{pasupat2015compositional}    & 4.7k   \\ \midrule 
    Total                                     &  -                & 409.1k  \\ \bottomrule
    \end{tabular}

    \vspace{-10pt}
    \label{tab:data}
    \end{table}

\subsection{Loss}
Since TextMonkey is trained to predict the next tokens like other LLMs, it only requires maximizing the likelihood of loss at training time.  
\def\bw{{\bf w}}
\def\bs{{\bf s}}
\def\bI{{\bf I}}
\def\bQ{{\bf Q}}
\begin{equation}
    \label{eq_objective}
    \mathcal{L} = {\rm max} \sum_{i=1}^{L} \log
     P(\tilde{{\bs}}_i | {\bI},{\bQ}, {\bs}_{1:i}),
\end{equation}
where $\textbf{I}$ is the input image, $\textbf{Q}$ is the question sequence, $\tilde{\textbf{s}}$ is the output sequence, $\textbf{s}$ is the input sequence, $ L$ is the length of the output sequence.

    \begin{table*}[hbtp]
    \centering
    \caption{Quantitative accuracy (\%) comparison of our model with existing large multimodal models (LMMs) on several benchmarks.  We fine-tune on TextMonkey with structured data shown in Sec.~\ref{subsec:dc}, resulting in TextMonkey\dag.}
    \resizebox{1\linewidth}{!}{\begin{tabular}{c|ccc|ccc|ccc|c}
    \toprule      \multirow{2}{*}{Method}
                 & \multicolumn{3}{c|}{Scene Text-Centric VQA}        & \multicolumn{3}{c|}{Document-Oriented VQA}                    & \multicolumn{3}{c|}{KIE} & \multirow{2}{*}{OCRBench}    \\
                 & STVQA & TextVQA & OCRVQA  & DocVQA & InfoVQA & ChartQA & FUNSD   & SROIE  & POIE  \\ \midrule
    BLIP2-OPT-6.7B~\cite{li2023blip2}   & 20.9  & 23.5    & 9.7            & 3.2    & 11.3           & 3.4                  & 0.2     & 0.1    & 0.3   & 235       \\
    mPLUG-Owl~\cite{ye2023mplug}    & 30.5  & 34.0      & 21.1           & 7.4    & 20.0             & 7.9           & 0.5     & 1.7    & 2.5   & 297     \\
    InstructBLIP~\cite{dai2023instructblip} & 27.4  & 29.1    & 41.3        & 4.5    & 16.4           & 5.3             & 0.2     & 0.6    & 1.0     & 276      \\
    LLaVAR~\cite{zhang2023llavar}       & 39.2  & 41.8    & 24.0             & 12.3   & 16.5           & 12.2              & 0.5     & 5.2    & 5.9   & 346       \\
    BLIVA~\cite{hu2023bliva}        & 32.1  & 33.3    & 50.7      & 5.8    & 23.6           & 8.7             & 0.2     & 0.7    & 2.1   & 291     \\
    mPLUG-Owl2~\cite{ye2023mplugowl2}   & 49.8  & 53.9    & 58.7        & 17.9   & 18.9           & 19.4            & 1.4     & 3.2    & 9.9   & 366       \\
    LLaVA1.5-7B~\cite{liu2023llava1.5}     & 38.1  & 38.7    & 58.1         & 8.5    & 14.7           & 9.3            & 0.2     & 1.7    & 2.5   & 297       \\
    TGDoc~\cite{wang2023TGDoc}   & 36.3 & 46.2    & 37.2     & 9.0           & 12.8                & 12.7       & 1.4    & 3.0   & 22.2   & -  \\
    UniDoc~\cite{feng2023unidoc}       & 35.2  & 46.2    & 36.8           & 7.7    & 14.7           & 10.9                    & 1.0       & 2.9    & 5.1   & -     \\
    DocPedia~\cite{feng2023docpedia}     & 45.5  & 60.2    & 57.2         & 47.1   & 15.2           & 46.9                 & {29.9}    & 21.4   & \underline{39.9}  & -       \\
    Monkey~\cite{li2023monkey}       & {54.7}  & \underline{64.3}    &{64.4}     & {50.1}   & {25.8}           & {54.0}            & 24.1    & {41.9}   & 19.9 & 514     \\ 
    InternVL~\cite{chen2024internvl}     & \textbf{62.2}  & 59.8    & 30.5           &28.7    & 23.6           & 45.6  & 6.5    & 26.4    & 25.9   &   517  \\
    InternLM-XComposer2~\cite{dong2024internlmxcomposer2}      & 59.6 & 62.2    & 49.6           &39.7    & \textbf{28.6}           & 51.6  & 15.3   & 34.2    & \textbf{49.3}  & 511 \\ \midrule
    
    TextMonkey   & \underline{61.8} & \textbf{65.9}    & \underline{71.3}     & \underline{64.3}          & \underline{28.2}                & \underline{58.2}       & \underline{32.3}    & \textbf{47.0}   & 27.9   & \textbf{561}  \\ 
    TextMonkey\dag & 61.2 &\underline{64.3} & \textbf{72.2} & \textbf{66.7} & \textbf{28.6} & \textbf{59.9} & \textbf{42.9} & \underline{46.2} & 32.0 & \underline{558} \\ \bottomrule
    \end{tabular}}
    \label{tab:result}
    \end{table*}

\section{Experiments}
\label{sec:experiments}
        \subsection{Implementation Details}
    \textbf{Model Configuration.}
    In our experiments, we utilized the well-trained Vit-BigG and LLM from Qwen-VL~\cite{bai2023qwen-vl}, which is a pre-trained large multimodal model. We configured the height and width ($H_v$, $W_v$) of the image inputs to 448 to align with the encoder specifications of Qwen-VL. Our image resampler is equipped with 256 learnable queries, and the token resampler's ratio (r) was set to 512 for images with a resolution of 896 and increased to 1024 for images with a resolution of 1344. To maximize training efficiency, our primary experimental focus was on using TextMonkey and evaluating outcomes at the 896 resolution setting.

    TextMonkey consists of a large language model with 7.7B parameters, an image resampler module with 90M parameters, a token resampler module with 13M, an encoder with 1.9B parameters, and Shifted Window Attention with 45M parameters. Overall, TextMonkey has a total of 9.7B parameters. 

    \textbf{Training.}
    During the training phase, we utilized the AdamW~\cite{adamw} optimizer, setting the learning rate to 1e-5 for the initial stage and reducing it to 5e-6 for the subsequent stage, while adopting a cosine learning rate schedule. The parameters $\beta_1$ and $\beta_2$ were configured to 0.9 and 0.95, respectively. A warmup period comprising 150 steps was incorporated, and we processed the data in batches of 128. To mitigate the risk of overfitting, we applied a weight decay factor of 0.1. The comprehensive training procedure spanned across 12 A800 days to complete one epoch.

    \textbf{Evaluation.}
    To facilitate a more equitable comparison with other approaches, we adopted the accuracy metric, where a response produced by our model is considered correct if it encompasses the ground truth. The selection of test datasets and the formulation of evaluation criteria were carried out in accordance with the methodology described in~\cite{liu2023hidden}. To ensure an even fairer comparison with other methods, we also performed supplementary evaluations on certain datasets utilizing their original metrics, such as F1 score and ANLS (Average Normalized Levenshtein Similarity).

\begin{table}[]
\centering
    \caption{Quantitative results on other document benchmarks. ``DF'' is an abbreviation for DeepForm. }
\label{tab:due}
\setlength{\tabcolsep}{4pt}
\begin{tabularx}{1.0\linewidth}{c|ccc|c|c|c}
\toprule
\multirow{2}{*}{Method} & \multicolumn{3}{c|}{Document} & Table & Chart& Scene \\
 & DocVQA & DF & KLC & WTQ & ChartQA& TextVQA \\ \midrule
Donut~\cite{kim2022donut} & 67.5 & \textbf{61.6} & 30.0 & 18.8 & 41.8&43.5 \\
Pix2Struct~\cite{lee2023pix2struct} & \underline{72.1} & - & - & - & 56.0&- \\
UReader~\cite{ye2023UReader} & 65.4 & 49.5 & 32.8 & 29.4 & 59.3&57.6 \\
Qwen-VL~\cite{bai2023qwen-vl} & 65.1 & 3.1 & 13.9 & 21.6 & \underline{65.7} &63.8\\
Monkey~\cite{li2023monkey} & 66.5 & 40.5 & \underline{33.9} & 25.3 & 65.1 &\underline{67.6}\\ \midrule
TextMonkey & 71.5 & \textbf{61.6} & \textbf{37.8} & \underline{30.6} & 65.5 & \textbf{68.0} \\
TextMonkey\dag & \textbf{73.0} & \underline{59.7} & \textbf{37.8} & \textbf{31.9} & \textbf{66.9} &65.6 \\ \bottomrule
\end{tabularx}

\end{table}
\subsection{Results}
\textbf{OCRBench Results.} We conduct a comparative analysis of our approach with recent large multimodal models. For our evaluation, we utilize three Scene Text-Centric VQA datasets: STVQA~\cite{STVQA}, TextVQA~\cite{singh2019towards}, and OCRVQA~\cite{mishra2019ocr}; three Document-Oriented VQA datasets: DocVQA~\cite{mathew2021docvqa}, InfoVQA~\cite{infovqa}, and ChartQA~\cite{masry2022chartqa}; and three Key Information Extraction (KIE) datasets: FUNSD~\cite{FUNSD}, SROIE~\cite{SROIE}, and POIE~\cite{kuang2023visual}. 
For a comprehensive assessment of performance, our evaluation includes OCRBench~\cite{liu2023hidden}, a recent benchmark specifically developed to evaluate the Optical Character Recognition (OCR) capabilities of Large Multimodal Models. OCRBench spans a wide range of text-related visual tasks, encompassing 29 datasets, and is designed to generate an overall score. 

\begin{figure*}[ht]
    \centering
    \includegraphics[width=0.93\linewidth]{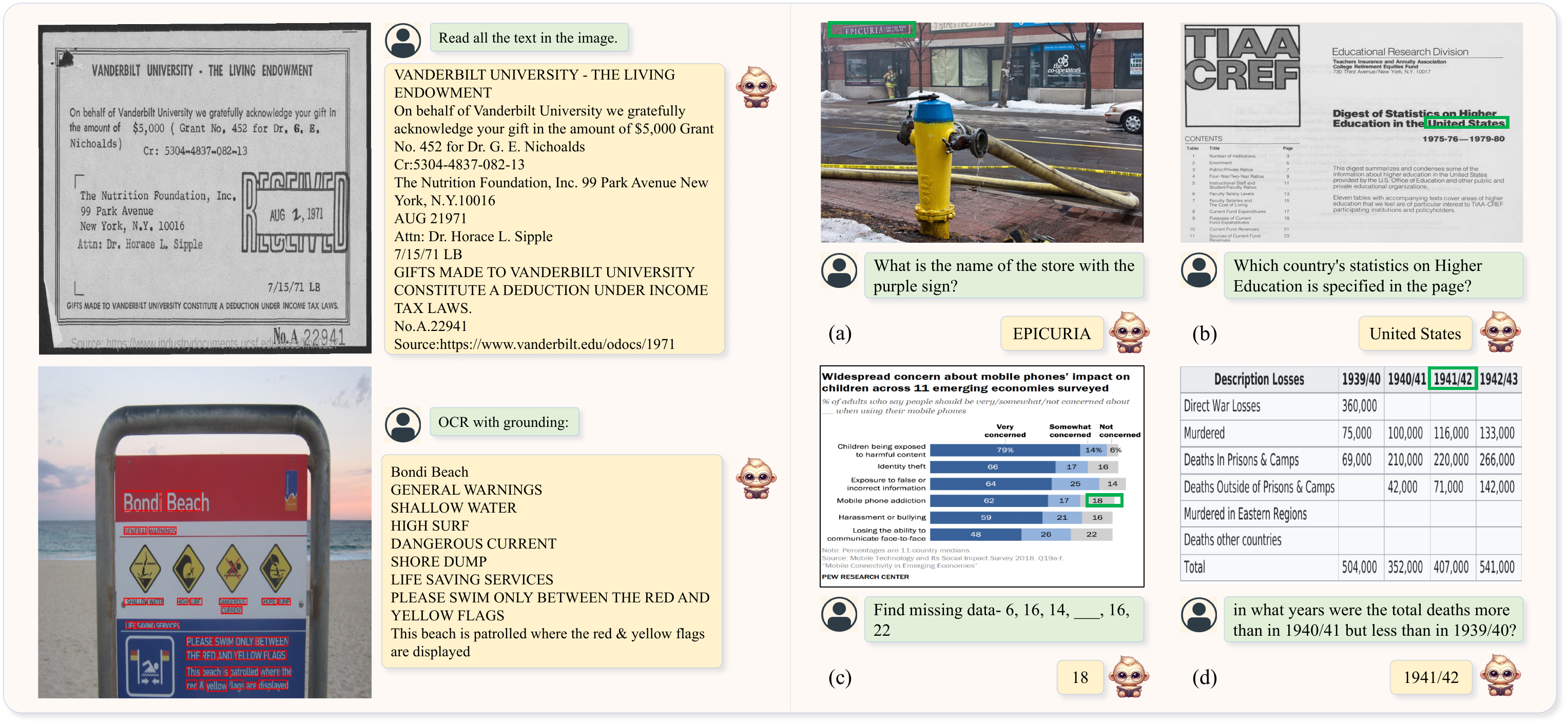}
    \caption{Visualization results of TextMonkey. The bounding boxes generated by the model are visualized in red. The location of the ground truths are highlighted with green boxes.}
    \label{fig:show}
\end{figure*}

As shown in Tab.~\ref{tab:result}, our model demonstrates superior performance compared to existing large multimodal models, particularly in scenarios where the text is dense and small. 
Our method inherently enhances many current evaluation datasets, resulting in average performance improvements with numerous baseline methods by 5.2\%, 6.9\%, and 2.8\% for Scene Text-Centric VQA, Document Oriented VQA and KIE, respectively.
TextMonkey can achieve 64.3\% on DocVQA and 58.2\% on ChartQA. Specifically, our model achieved a score of 561 on OCRBench. 
The performance on both two challenging downstream tasks and OCRBench demonstrates its effectiveness in text-related tasks. We have found that our model tends to provide numerical answers without units, which results in a performance decrease on POIE. 

\begin{table}[]
\centering

\caption{Quantitative accuracy of text spotting. The ``Total-Text'' and ``CTW1500'' datasets do not use a specific vocabulary for evaluation, while the ``ICDAR 2015'' dataset uses a general vocabulary for evaluation of other models. Note TTS only uses synthetic location data. TextMonkey is not fine-tuned by the downstream text spotting datasets without any vocabulary. }
\begin{tabular}{c|cc|cc|cc}
\toprule
 \multirow{2}{*}{Method} & \multicolumn{2}{c|}{Total-Text~\cite{ch2017total}} & \multicolumn{2}{c|}{CTW1500~\cite{liu2019curved}} & \multicolumn{2}{c}{ICDAR 2015~\cite{karatzas2015icdar}} \\
 & Trans & Pos & Trans & Pos & Trans & Pos \\ \midrule
TOSS~\cite{tang2022you} & 61.5 & 65.1 & 51.4 & 54.2 & 47.1 & 52.4 \\
TTS~\cite{kittenplon2022towards} & - & 75.1 & - & - & - & 70.1 \\
SPTS v2~\cite{liu2023spts} & 64.7 & \textbf{75.5} & 55.4 & 63.6 & 55.6 & \textbf{72.6} \\ \midrule
TextMonkey & \textbf{78.2} & 61.4 & \textbf{63.2} & 57.5 & \textbf{66.9} & 45.1 \\
\bottomrule
\end{tabular}
\label{tab:spot}
\end{table}

\textbf{Document Benchmarks results.} To further compare and assess the capabilities of our method,
we conduct tests on additional datasets utilizing the specific evaluation metric provided in their paper: F1-score for Deepform and KLC, accuracy for WTQ, relaxed accuracy measure for ChartQA, ANLS for DocVQA, and VQA score for TextVQA. 

The results, shown in Tab.~\ref{tab:due}, indicate that our model leads in performance on these datasets, outperforming other models. 
Across different domains, TextMonkey achieves a score of 
 71.5 in DocVQA, 30.6 in WTQ, 65.5 in ChartQA and 68.0 in TextVQA. 
It shows our model's capability to handle documents, tables, charts, and scene text.

\textbf{Text spotting results.} To show the extensive capabilities of our model, we assessed its 
performance on text spotting datasets without finetuning, as detailed in Tab.~\ref{tab:spot}. Given our model's focus on identifying complete text passages, we segmented the predicted content into individual words for analysis. We employed two evaluation methodologies to evaluate our model's performance. In the ``Trans'' mode, text is considered correct if the answer contains this word. 
Conversely, the ``Pos'' mode requires the consideration of positional information in accordance with previous methods~\cite{liu2021abcnetv2}.
For both metrics, due to granularity issues of the output (TextMonkey often produces an integrated paragraph while others only produce desired words), the metric can not strictly follow the evaluation setup; however, both should be quite similar, as both the error and correct situations match in calculations. 

To maintain TextMonkey's consistent performance, we refrained from fine-tuning it with downstream text spotting data, unlike other methods that were optimized for either the ``Trans'' or ``Pos'' metrics. Our results reveal that, for the ``Trans'' metric, TextMonkey outperformed SPTS v2 by a margin of 10.9\%. Regarding the ``Pos'' metric, it demonstrated competent text spotting capabilities, showing its ability in understanding both text content and spatial positioning.

    \subsection{Visualization}
    We conduct a qualitative evaluation of TextMonkey across various scenarios, including natural scenes and document images. As shown in the left part of Fig.~\ref{fig:show}, TextMonkey accurately locates and identifies text in both scene and document images. Besides, natural images in Fig.~\ref{fig:show} (a), documents in Fig.~\ref{fig:show} (b), charts in  Fig.~\ref{fig:show} (c), and tables in Fig.~\ref{fig:show} (d) exemplify TextMonkey's adeptness at discerning and interpreting visual and textual information within a wide range of scenarios.
    Overall, TextMonkey's performance across diverse scenarios demonstrates its effectiveness in perceiving and comprehending textual information in various visual contexts.

\begin{table}[]
\centering
\caption{Ablation study on zero initialization.}
\begin{tabular}{c|cccc}
\toprule
Zero Initialization & SROIE & DocVQA & TextVQA & ChartVQA \\ \midrule
$\times$ & 46.8 & 64.1 & 65.7 & 57.6 \\
$\checkmark$ & \textbf{47.0} & \textbf{64.3} & \textbf{65.9} & \textbf{58.2} \\ \bottomrule
\end{tabular}
\label{tab:lora}
\end{table}
 \begin{table}[t]
\centering
\caption{Ablation study on different components. ``W-Attn" means Shifted Window Attention, ``T-Resampler" means Token Resampler. }
\begin{tabular}{cc|ccc}
\toprule
W-Attn           & T-Resampler        & SROIE       & DocVQA       & TextVQA      \\ \midrule
$\times$    & $\times$ &    45.9         &             62.6 &         62.4     \\
$\checkmark$   & $\times$ & 46.0       & 64.1        & 64.8        \\
$\checkmark$    & $\checkmark$ & \textbf{47.0}       & \textbf{64.3}        & \textbf{65.9}        \\ \bottomrule
\end{tabular}

\label{tab:componet}
\end{table}

\begin{table}[t]
\centering
\caption{Effectiveness of the strategy of Token Resampler. }
\begin{tabular}{l|ccc}
\toprule
Method                 & SROIE       & DocVQA       & TextVQA      \\ \midrule
w/o token filter      & 32.9       & 46.7        & 59.5        \\
w/o resampler          &   44.9          &           63.5   &    62.5          \\

w unsorted token filter & 46.8       & 62.1        & 64.2         \\
ours                   & \textbf{47.0}       & \textbf{64.3}        & \textbf{65.9}        \\ \bottomrule
\end{tabular}
\label{tab:token}
\end{table}
   
   \subsection{Ablation Study}
   \textbf{Ablation study on zero initialization.} Since CLIP is already pretrained, it is advisable to avoid drastic changes in features during the early training stages. As shown in Tab.~\ref{tab:lora}, incorporating this zero initialization method can yield 0.6\% performance gain on ChartQA.
   
    \textbf{Ablation study on different components.} As shown in Tab.~\ref{tab:componet},
    by introducing cross-window connections, we achieved an improvement of 0.1\% on SROIE, 1.5\% on DocVQA, and 2.4\% on TextVQA. It can be observed that cross-window connections partially compensate for the discontinuity caused by chunking and contribute to a better understanding of the images.  Based on the Token Resampler, our method demonstrates better performance, achieving 1.0\%, 0.2\%, and 1.1\% performance gain on the SROIE, DocVQA, and TextVQA. This suggests that our approach effectively preserves essential information while eliminating redundant tokens, thereby simplifying the learning process for the model.
    
    \textbf{Ablation study on strategies of reducing token length.} 
    As demonstrated in Tab.~\ref{tab:token}, substituting important tokens with random ones (without token filter) leads to an average decline in performance by roughly 12.7\%. This decline is attributed to the increased complexity of optimizing random queries, which necessitates more datasets to achieve a generalized representation compared to utilizing significant tokens. Solely focusing on pivotal features (without resampler) and directly eliminating features incurs a loss of some information, showing a decrease in performance, such as a 2.1\% drop in SROIE. Additionally, neglecting the order of tokens (with unsorted token filter) does not markedly impair performance, owing to the language model's inherent ability to organize unordered tokens. Nevertheless, the lack of token order can still lead to decrease, especially evident in the result of DocVQA, with a 2.2\% decrease in performance.

    \textbf{Interaction between input resolution and the number of tokens remained.} As shown in Tab.~\ref{tab:res}, 
    Directly increasing the resolution without compressing tokens can actually lead to consistent worse performance, especially with a decrease of 9.2\% performance in DocVQA. We speculate that the increase in resolution results in a significant increase in redundant tokens, making it more difficult to find crucial information in our setting. 
    Therefore, compressing tokens reasonably can lead to higher performance.
    Considering the sparsity of information in large-sized images, it is also necessary to consider selecting an appropriate value of ``r'' for different input resolutions.
    Besides, increasing the input resolution brings benefits to the dataset, which contains many large-sized images, with 0.2\% performance gain for DocVQA and 3.2\% performance gain for InfoVQA. However, for datasets like TextVQA and SROIE,  which contain much smaller images, increasing the input resolution directly does not yield any gains.

\begin{table}[t]

\caption{Interaction between resolution and the number of tokens remained ``r''. ``-'' in ``r'' means do not use token resampler and keep all the remaining tokens. }
\label{tab:res}
\begin{tabular}{cc|cccc}
\toprule
Resolution & r & SROIE & DocVQA & TextVQA & InfoVQA \\ \midrule
896 & - & 46.0 & 64.1 & 64.8 & 29.1 \\
896 & 256 & \textbf{47.0} & 60.9 & 65.2 & 25.9 \\
896 & 512 & \textbf{47.0} & 64.3 & \textbf{65.9} & 28.2 \\
1344 & - & 42.9 & 54.9 & 62.5 & 28.9  \\
1344 & 512 & 44.9 & 59.7 & 64.2 &28.0  \\
1344 & 1024 & 46.0 & \textbf{64.5} & 65.1 & \textbf{31.4} \\ \bottomrule
\end{tabular}
\end{table}

    \begin{figure}[ht]
        \centering
    \includegraphics[width=0.93\linewidth]{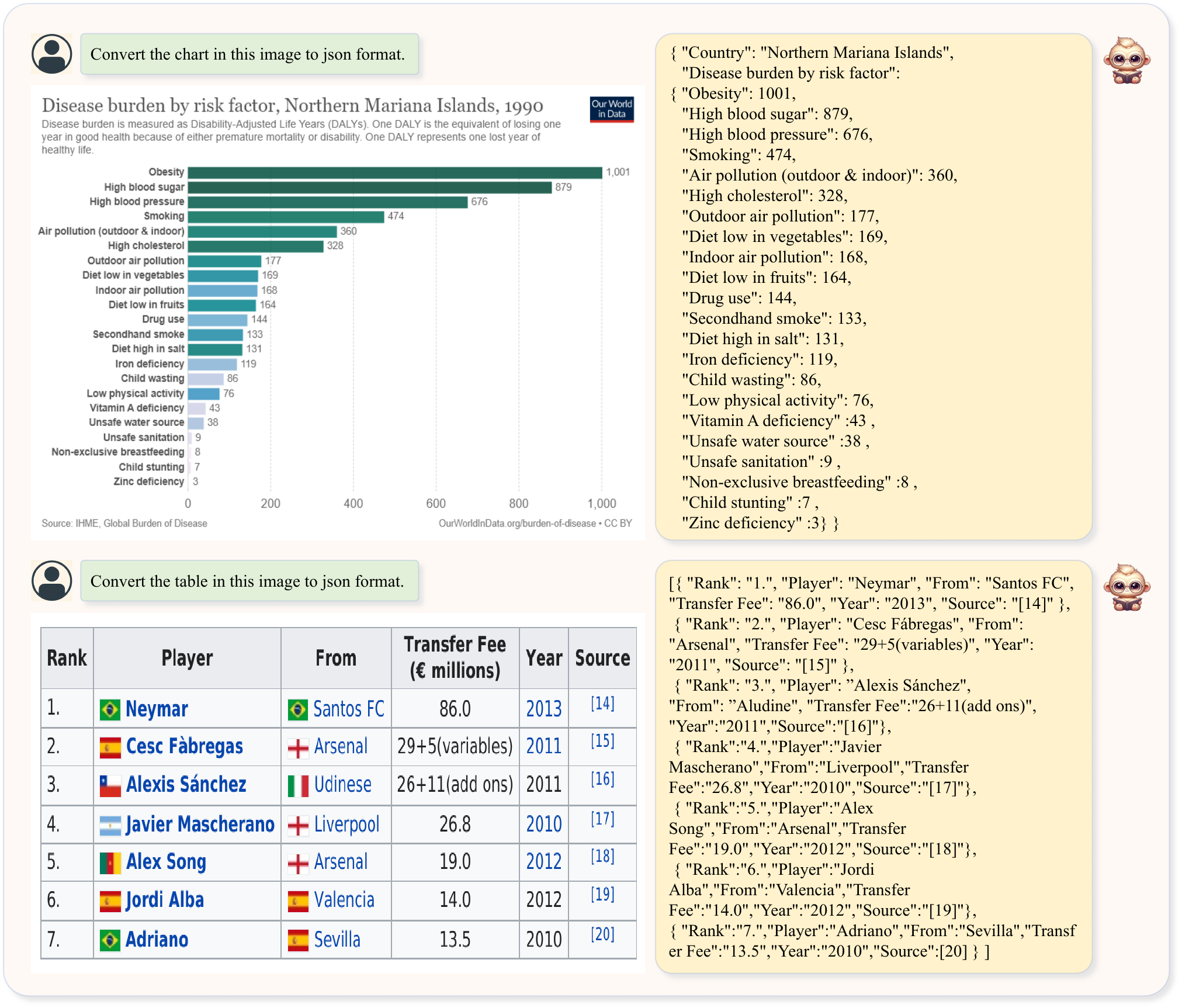}
        \caption{Examples of structuralization of chart and table using TextMonket\dag.}
        \label{fig:struct}
    \end{figure}

    \subsection{Structuralization}
    The structuralization of charts and tables holds substantial practical value. Structured charts and tables present data in a clear format, and by extracting structural information from images, computers can accurately parse and extract the data. This makes data analysis, statistics, and modeling more efficient and precise. It also helps reduce the complexity of information and improves its comprehensibility. As depicted in Fig.~\ref{fig:struct}, our model is capable of structuring charts and tables into JSON format, demonstrating its potential for downstream applications. According to Tab.~\ref{tab:due}, {TextMonkey\dag}  exhibits a performance improvement of 1.3\% and 1.4\% on tables and charts, respectively. This underscores that high-quality data not only enables the model's structuralization capabilities but also amplifies the effectiveness of the related benchmarks. However, it is worth noting that this type of data will primarily benefit the data within its own domain, thus leading to a performance decrease for cross-domain TextVQA.
    
    \begin{figure*}[ht]
    \centering
    \includegraphics[width=0.93\linewidth]{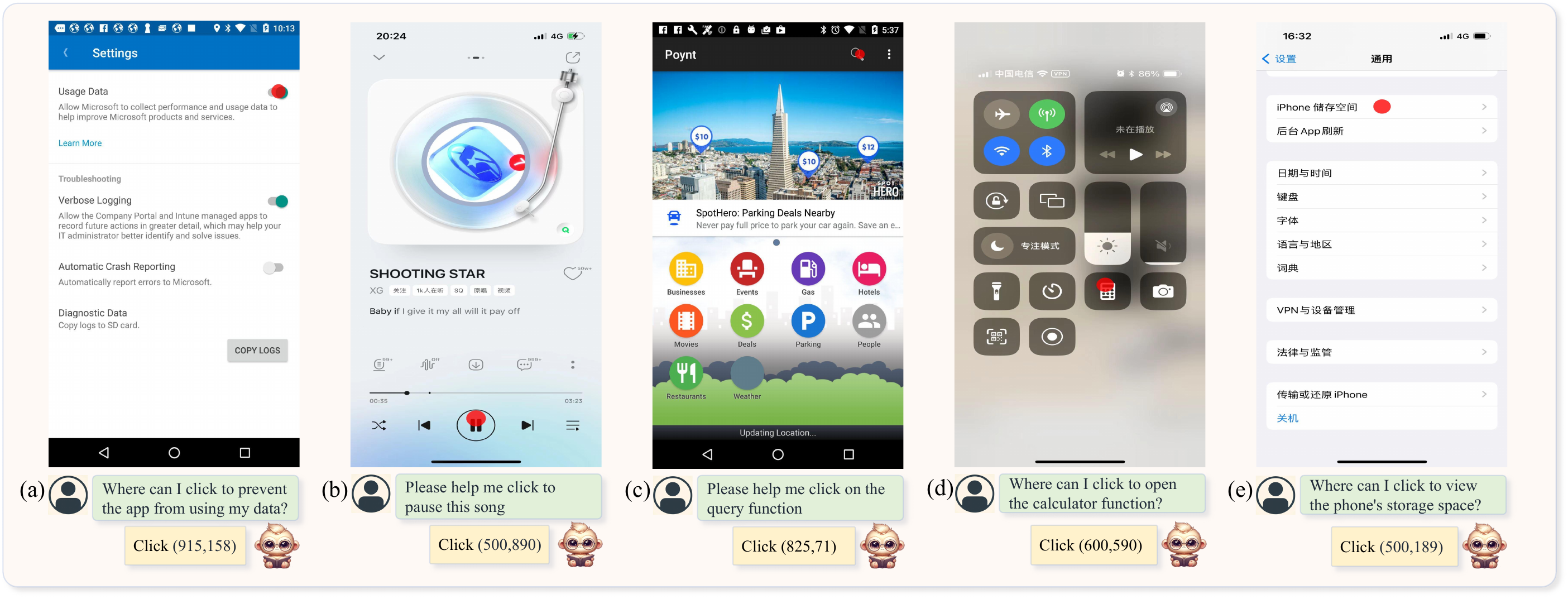}
    \caption{Visualization results of fine-tuned TextMonkey for Apps. The clicking results generated by the model are visualized in red point. To better simulate human behavior, we have magnified the points into circles.}
    \label{fig:agent}
\end{figure*}
    \subsection{App Agent}
   Recently, there has been a lot of attention on using LMMs for the task of acting as agents for smartphone applications ~\cite{yang2023appagent,wang2024mobile,niu2024screenagent}. Unlike existing intelligent phone assistants like Siri, which operate through system back-end access and function calls, this agent interacts with smartphone applications in a human-like manner, using low-level operations such as clicking and swiping on the graphical user interface (GUI). It eliminates the need for system back-end access, enhancing security and privacy as the agent does not require deep system integration. The GUI primarily consists of icons and text, and we explore the feasibility of TextMonkey on this aspect. We transformed 15k user click data from the Rico~\cite{deka2017rico} dataset and performed downstream fine-tuning using TextMonkey. As qualitatively shown in Fig.~\ref{fig:agent}, our model is able to understand user intent and click on the corresponding icons, which suggests the potential of the model to serve as an app agent by using downstream data.

  \begin{table}[]
\centering
\caption{Effect of incorporating the position of answer}
\begin{tabular}{l|cccc}
\toprule
Method & DocVQA & SROIE & ChartQA & InfoVQA \\ \midrule
w position & \textbf{64.5} & \textbf{47.2} & 57.8 & 27.7 \\
w/o position & 64.3 & 47.0 & \textbf{58.2} & \textbf{28.2} \\ \bottomrule
\end{tabular}
\label{tab:pos}
\end{table}

    \begin{figure}[ht]
        \centering
    \includegraphics[width=0.93\linewidth]{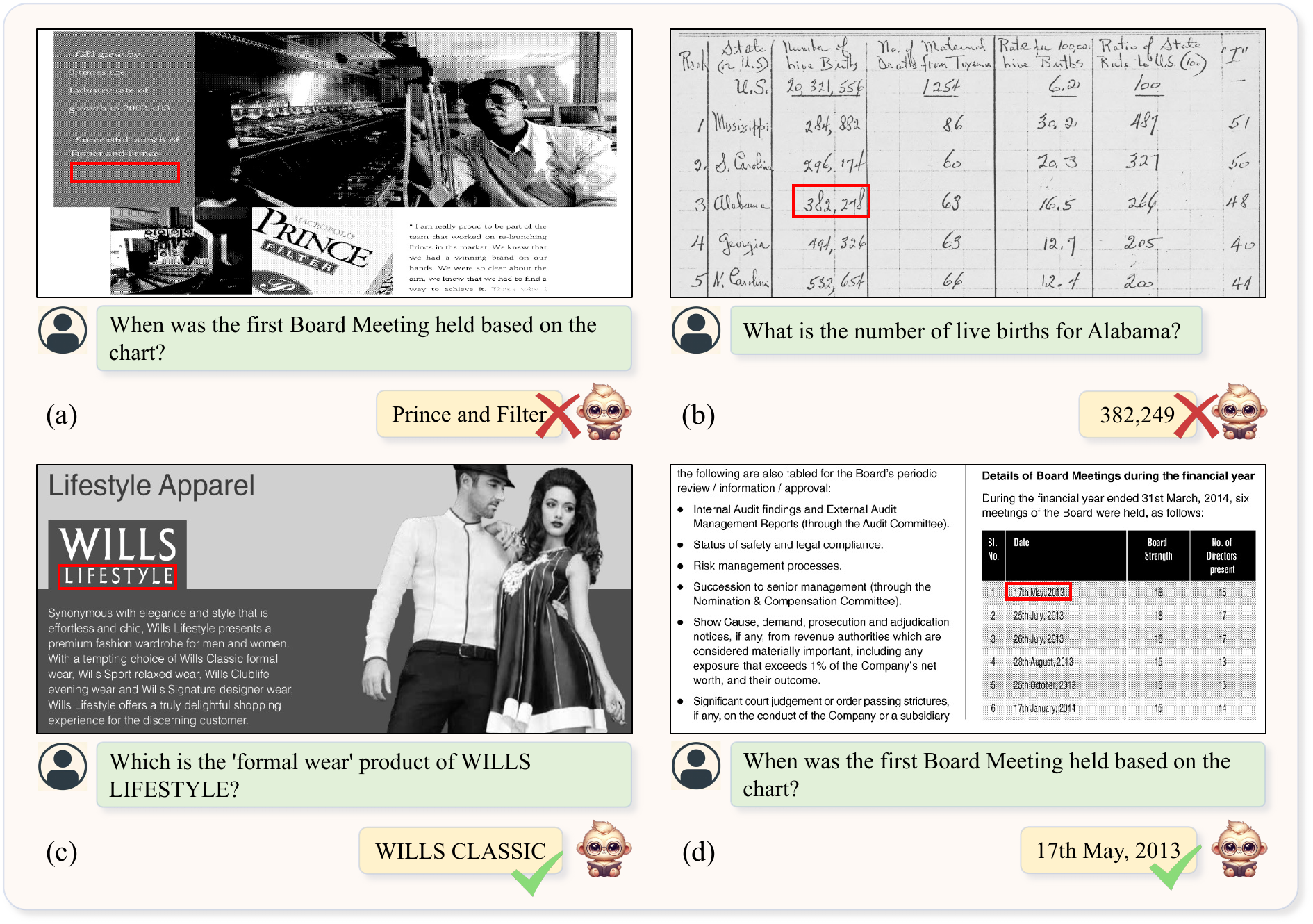}
        \caption{Examples of grounding the position of the answer.}
        \label{fig:dis}
    \end{figure}

    \section{Discussion}
    \subsection{Interpretability}
    By examining the grounding information, we can identify the reasons behind the model's  errors, 
    enhancing a better understanding of the model's behavior. As shown in Fig.~\ref{fig:dis} (a), we ground to the white region, indicating that the model might be engaging in hallucination. We correctly identify the location but recognize it wrongly in Fig.~\ref{fig:dis} (b). Fig.~\ref{fig:dis} (c) highlights a scenario where the model grounds to incorrect text but still provides the correct answer. This could imply that there is some level of randomness or uncertainty in the model's responses at this point.  In  Fig.~\ref{fig:dis} (d), the alignment between the position and text indicates that the model is more confident in its predictions. Therefore, based on these analyses, we can gain a better understanding of the model's behavior and have a better awareness of the model's hallucination, thus reducing the model's hallucination.

   \subsection{Chain-of-Thought} We also conduct experiments on several datasets and observe inconsistent improvements if we require a model to provide the answer's position, as shown in Tab.~\ref{tab:pos}.    In datasets where the majority of answers are based on information within the images, such as DocVQA and SROIE, there is a noticeable benefit in requiring the model to provide the answer's position. However, for datasets that involve reasoning tasks, such as ChartQA and InfoVQA, where questions require comparisons or quantitative analysis (e.g., "How much more is A than B?"), demanding positional answers can actually result in a detrimental effect.
   Upon further examination of the wrong answer, we consider that the requirement of grounding might have partially affected certain reasoning needs. 
    Hence, it is essential to consider the nature of the dataset and the type of questions being asked when deciding whether to impose the requirement of positional answers.
    
   Additionally, we believe that automating the process of constructing a thinking chain~\cite{wei2022chain} in subsequent steps could be a promising direction for future research. By developing mechanisms to generate a coherent chain of reasoning automatically, we can potentially enhance the overall performance and reasoning capabilities of our models.
    \begin{table}[]
    \centering
\caption{Comparison with different shapes of bounding
box.}
\begin{tabular}{c|cccc}
\toprule
Representation & SROIE & DocVQA & TextVQA & ChartVQA \\ \midrule
Polygon & 47.2 & 64.0 & 65.7 & 57.9 \\
Rect & 47.0 & 64.3 & 65.9 & 58.2 \\
Point & \textbf{47.9} & \textbf{65.0} & \textbf{66.0} & \textbf{58.3} \\ \bottomrule
\end{tabular}
\label{tab:bbox}
\end{table}

    \subsection{Comparison Between Different  Representations of Position}
    Recently, some methods~\cite{liu2023spts} have used points to represent positions instead of rectangles and polygons.  Firstly, intuitively, the cost of generating points during inference would be lower compared to generating rectangles and polygons, as generating Nx points is required for other forms of bounding boxes.
    We aim to further investigate and experimentally validate which form is more suitable for LMMs to learn.
    To maintain strict consistency in our experiments, we only applied transformations to the data while keeping the other training hyperparameters the same. For the points, we selected the center points of the bounding boxes that were the most meaningful.

    As demonstrated in Table~\ref{tab:bbox}, employing points as visual cues significantly enhances performance over rectangles. In the case of Docvqa, there was an improvement of 0.7\%, while for SROIE, the enhancement reached 0.9\%. Furthermore, rectangles often surpass polygons in performance. This might be attributed to the previously discussed issue that redundant image tokens could increase the complexity of the model's learning process. Similarly, extensive position representations might face comparable obstacles. Given these considerations, along with the associated inference costs, utilizing points as representations can be a viable strategy for appropriate tasks.

\section{Conclusion}
This paper introduces TextMonkey to address the challenges associated with text-heavy tasks such as document question answering and fine-grained text analysis.
We adopt Shifted Window Attention with zero initialization to help establish relationships while increasing input resolutions using a sliding window. Increasing the resolution simultaneously increases the number of tokens. Through analyzing the redundancy of tokens, our proposed Token Resampler effectively reduces the number of tokens. 
Furthermore, by engaging in multiple text-oriented tasks simultaneously, TextMonkey enhances its perception and understanding of spatial relationships, leading to improved interpretability and support for clicking screen-shots. By comparing our model with various LMMs, our model achieved excellent results on multiple benchmarks. It is worth mentioning that we also find that directly increasing the input resolution does not always lead to improvements, particularly for much smaller images. This underscores the necessity of creating an efficient method for scaling resolution in documents where size changes can be dramatic.

\ifCLASSOPTIONcaptionsoff
  \newpage
\fi

{
\bibliographystyle{IEEEtran}
 \bibliography{refer}

}

\vfill 
\end{document}